# Architecture-Based Classification of Plant Leaf Images

Mahmoud Sadeghi[1], Ali Zakerolhosseini[2], Ali Sonboli[3]

## Abstract

Plant classification and identification is an important and difficult task. In this paper, a systematic approach for extracting the leaf architecture characters from captured digital images is proposed. The input image is first pre-processed to be prepared for feature extraction. In the second stage, different architectural features are extracted and mapped to semantic botanical terms. Lastly, we propose a method for classifying leaf images based on these features. Compared with previous studies, the proposed method combines extracted features of an image with specific knowledge of leaf architecture in the domain of botany to provide a comprehensive framework for both computer engineers and botanist. Finally, based on the proposed method, experiments on the classification of the images from ImagerCLEF 2012 dataset has been performed and results are presented.

[1] Mahmoud Sadeghi
E-mail: m_sadeghi@alum.sharif.edu

[2] Ali Zakerolhosseini
Department of Electrical and Computer Engineering
Shahid Beheshti University,
1983963113, Evin, Tehran, Iran
E-mail: a-zaker@sbu.ac.ir

[3] Ali Sonboli
Department of Biology
Medicinal Plants and Drugs Research Institute
Shahid Beheshti University,
1983963113, Evin, Tehran, Iran
E-mail: a-sonboli@sbu.ac.ir

# 1. Introduction

Plant identification is a demanding and challenging task, especially due to the vast number of plant species. Traditionally, plant taxonomy has been carried out only by expert botanists. But, in the past decade, the expansive development and ubiquity of technologies such as digital imaging and portable computing propel this task more toward automation.

Professional botanists use different parts of a plant for identifying its species: Leaves, flowers, fruits, seeds and bark. Among these, leaves are of more prominent importance. They are available for much of the year and they contain a great deal of information for a skilled botanist to examine. Moreover, considering the leaves' surface as mostly flat and two dimensional, they can be analyzed using images.

Plant leaves have two main characters that are often used in species identification: *shape* and *vein pattern*. Each of these can be analyzed separately or combined (fused) in classification tasks. However, the leaf shape characters are the more reliable feature to study due to the difficulty of the observing vein patterns and their sensitivity to the image capturing environments. Nevertheless, extraction and analysis of the leaf vein patterns is the base for many researches [1, 2, 3, 4] .

Early works on systematic shape based image retrieval are of Wang *et al* [5, 6] who used different sets of features such as centroid-contour distance (CCD), angle code histogram (ACH) and Fourier descriptors for shape representation and later fuse them together using fuzzy integral. Du et *el* [7] proposed move median centers (MMC) for classifying leaves based on extracted morphology features like aspect ratio, eccentricity and rectangularity. They showed promising results on an in-house dataset. Elliptic Fourier transform (EFD) is the base of the work of Neto *et al* [8] who applied it for leaf shape analysis. Chen *et al* [9] proposed using velocity representation method for describing the leaf contour shape. Mokhtarian and Abbasi [10] used curvature scale space (CSS) to represent leaf shapes and applied it to leaf classification with self-intersection. Moreover, plant leaf modeling methods such as diffusion limited aggregation (DLA) [11], standard cutting models (SCMs) [12], neighborhood rough set [13] and Gabor filters [14, 15] have also been studied.

Although many of the above mentioned methods yield encouraging classification results, they employ features defined solely in computer science. In botany, the leaf "architecture" and its characters [16] are described semantically and used for species identification. But little attempts have been made to formulate these specifications and apply them to classification tasks. This approach is carried out to some extent by Fu *et al* [17]. However, they didn't offer any classification or plant identification results.

In this study, we propose algorithmic formulations for describing taxonomical features and classification methods for plant species identification. We believe this approach is interdisciplinary and can be used by computer scientists and botanist alike. Our proposed method consists of three main steps: (1) Leaf image preprocessing, (2) Architecture-based features extraction and (3) Classification and species identification.

The rest of this paper is organized as follows: in section 2, we explain how the input leaf image is preprocessed. The feature extraction and semantic rules are described in section 3. In section 4, we propose a method for plant species identification and the results are presented in Section 4. Finally, we conclude our work and suggest the future works in Section 5.

## 2. Leaf image preprocessing

In this section, we present the preprocessing steps performed on the input image to extract the shape of the leaf and provide the convenient input for feature extraction.

### 2.1. Segmentation

Since we want to study only the shape of a leaf, the leaf color or its background are of no use to our method. We use Otsu segmentation algorithm [18] on the color image (Fig.1 (a)) to convert it to a binary image. This algorithm finds the optimum threshold to segment the image pixels to two classes (i.e. foreground and background), which is suitable here, as we need to separate the leaf from its background and ignore its color attribute. Fig.1 (b) shows the result of applying Otsu's algorithm on a leaf image

## 2.2. Petiole removal

Although an important part of a plant leaf in botany, the petiole is often disregarded in leaf character studies and almost all of the main features are focused to the leaf's *lamina* (the body). Therefore, it is preferable that the petiole be omitted from the leaf image.

A common method to achieve this is to utilize the narrowness of the petiole in comparison to the width of the leaf for identifying the petiole and removing it. This method, while effective in some cases, poses difficulties in the cases of leaves with basal extension (discussed in section 3.4), not to mention the heuristic task of finding an adequate threshold. Instead, we propose applying morphological operations for petiole removal. First, the binary image of the leaf is eroded and then dilated using circular element with diameter of 5% of the image width. Then, the result is subtracted from the original image, yielding an image containing only the petiole. Removing this from the original image will result in the image of lamina (Fig.1 (c)).

Also at this stage, the (approximate) insertion points of the petiole are determined from the extracted petiole image (we assume the petiole is always at the bottom of the leaf). We identify 2 connection points at two sides of the top of the petiole for use in later stages of classification. This helps in checking for the basal extension by comparing it them the lowest point of the lamina: If the connections points are not the same as the lowers point of the lamina, the leaf has a basal extension and the base angle is identified as *reflex* (refer to section 3.7).

## 2.3. Apex terminal point identification

The apex terminal point (here referred to as simply the apex point) is the top point of the midvein, where it meets the margin (Fig.1 (d)). It plays an important role in finding the shape features of a leaf. Without the availability of the midvein trend, finding the apex terminal accurately can lead to better feature extraction, even features not directly related to the apex. The apex point is either the top point of the leaf (aligned with petiole at the bottom), or it is at the base of a concavity on the top of the leaf. These two types are identified by checking for a minimum point in the edge of top 25% of the lamina. If the minimum point is one of the two bottom points, the apex point is at the top (indicated by a circle in Fig.1 (d)). In the case of the top of the leaf being concave, the leaf has an *apical extension* resulting in apex angle being *reflex* (section 3.7).

**2.4. Boundary extraction**

Having the binary image of the lamina (referred to as simply "the leaf" along the paper), its boundary can easily be extracted using common boundary extraction techniques [19]. Fig.1 (e) shows the extracted boundary of a leaf.

**2.5. Geometric representation**

Although the extracted boundary should suffice in many feature extraction tasks, the image coordination of the boundary points can be inconvenient to use for some geometric calculations. Thus, we convert it to the Cartesian coordination for easier perception and computation. Also, having the apex point from the previous section and assuming the leaf's midvein has no significant curve, we rotate the boundary so an estimated midvein (the line between apex point and the middle of petiole connection points) will be vertical (Fig.1 (f)).

The geometric representation of the leaf boundary is the main input for the classification stage. From here forth, it is simply referred to as "the boundary".

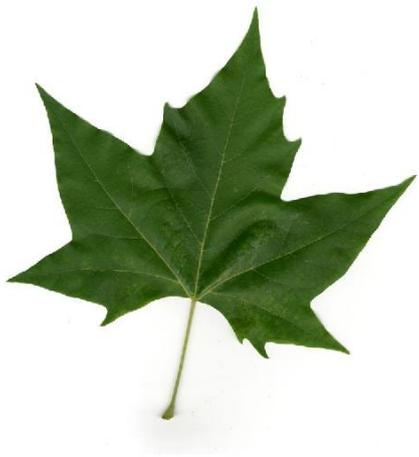

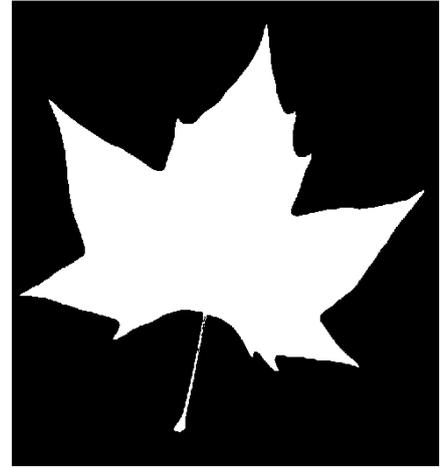

(a)　　　　　　　　　　　　　　　(b)

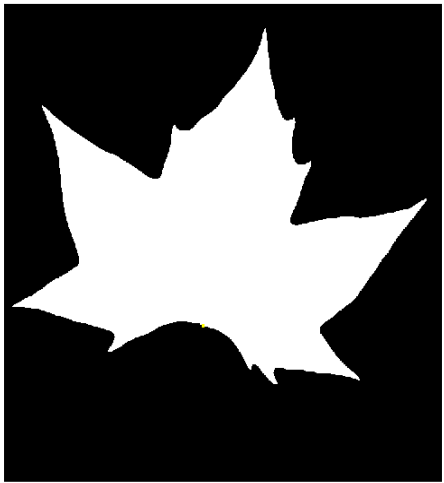

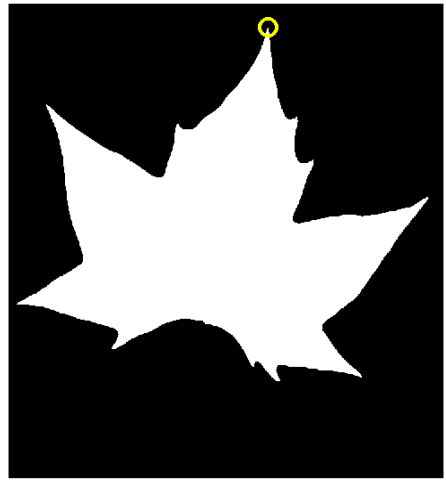

(c)　　　　　　　　　　　　　　　(d)

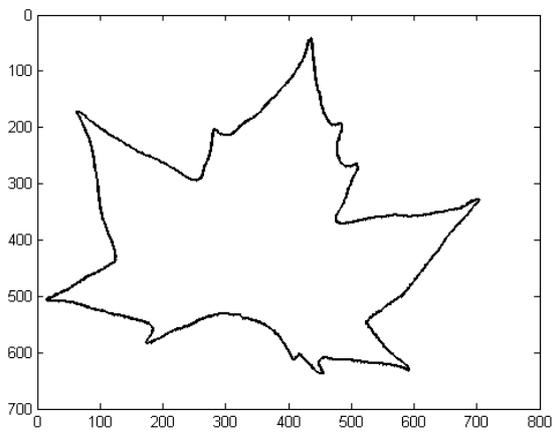

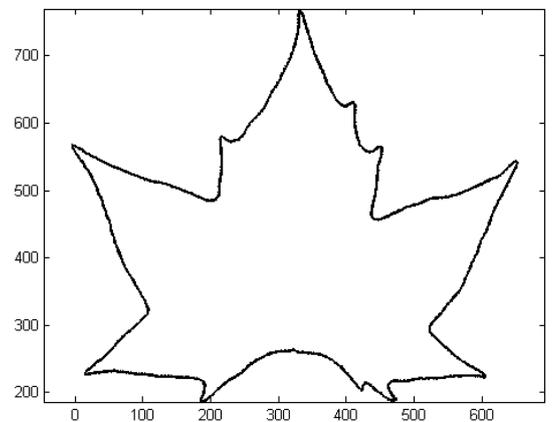

(e)　　　　　　　　　　　　　　　(f)

**Figure 1** Preprocessing steps of a plant leaf image (a) The leaf image; (b) binary image; (c) leaf image with petiole removed; (d) apex point identification; (e) extracted boundary; (f) geometric representation of the boundary

## 3. Architecture-based features extraction and semantic rules

As discussed in introduction, it is beneficial to formulate architectural features and map these calculations to sematic attributes used by botanists.

To accomplish this task, the quantitative definitions alongside the terminology presented in [16] are employed as the source for feature extraction and categorization. In this section, we present the methods used for extracting each semantic feature and how the leaf is categorized regarding that attribute.

### 3.1. Leaf organization

Laminar organization is the highest level feature which can be recognized almost at the first glance. Leaves are organized into two categories: (1) Simple and (2) Compound. Simple leaves consist of a single lamina attached to a single petiole. Compound leaves consist of two or more leaflets attached to a single rachis. Fig.2 shows examples of a simple and a compound leaf.

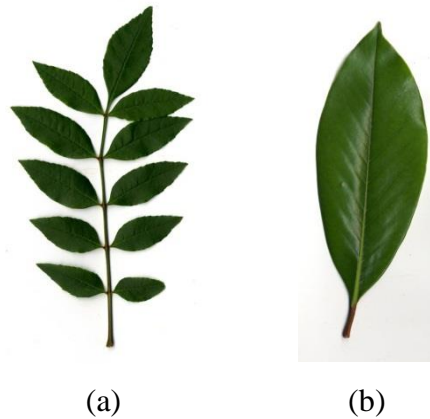

(a)     (b)

**Figure 2** an example of (a) compound and (b) simple leaves

We use morphological operations for separating simple and compound leaves. The binary image of the leaf (output of stage 1.1) is eroded with a circular element with 3% of the width of the image in diameter. In an image consisting of a single leaf, only one shape will remain. But, in case of compound leaves, this operation leaves at least 2 disjoint objects. Thus, by calculating the number of connected components using *flood-fill* algorithm, the categorization can be easily achieved (Fig. 3).

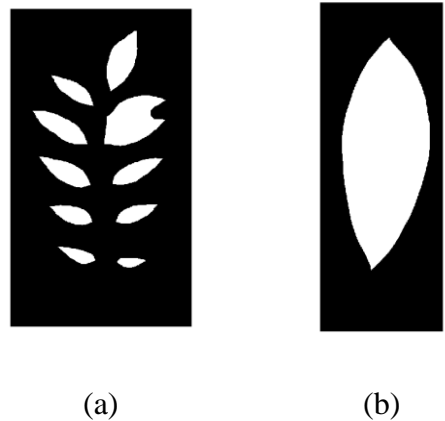

(a)　　　　　　　　(b)

**Figure 3** Results of the erosion operation on the leaves of Fig.2 (a) Compound leaf - 10 disjoint objects (b) Simple leaf - 1 object

### 3.2. Laminar characters

In [16], two characters for the leaf lamina is defined: *Laminar L:W ratio* and *laminar shape.* The L:W ratio indicates the ratio of laminar length to its maximum width (perpendicular to the axis of the midvein) (Fig.4). In our geometric representation of the leaf, this feature can be stated mathematically as $\frac{Y_{max}-Y_{min}}{width_{max}}$.

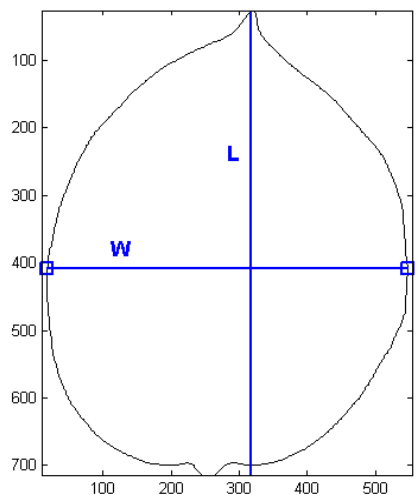

**Figure 4** L:W ratio of the leaf

The Laminar shape is basically determined by where the widest point is located. Based on this location and L:W ratio, the leaves are categorized into 5 classes [22]:
(1) *Elliptic* - The widest part of the leaf is in the middle one-fifth

(2) *Obovate* - The widest part of the leaf is in the distal two-fifths

(3) *Ovate* - The widest part of the leaf is in the proximal two-fifths

(4) *Oblong* - The opposite margins are roughly parallel for at least the middle one-third of the leaf

(5) *Linear* - The L:W ratio of the leaf is ≥ 10:1, regardless of the position of the widest part of the leaf

Classifying the laminar shape is straightforward by just comparing the calculated values to the thresholds expressed in the Table.1. In the case of the *Oblong* leaves, we premise 2.5% of the image width as the threshold of equality of the parallel lines. Fig.5 shows one leaf classified into each of the above classes (the green lines depict the thresholds).

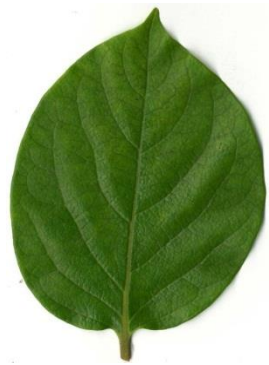 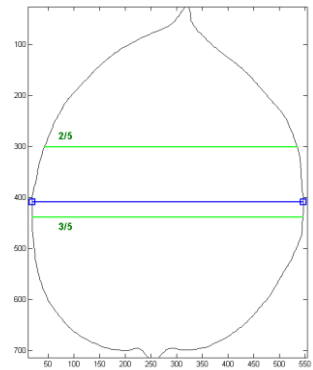

(a)

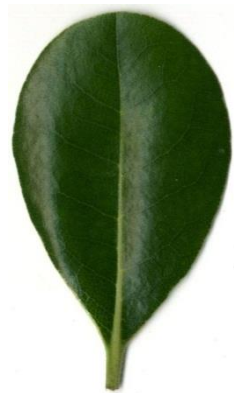 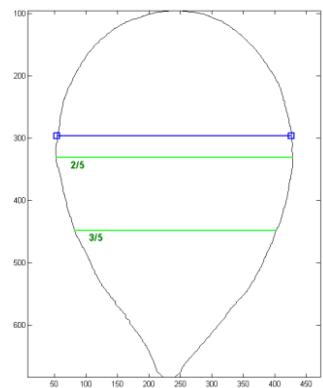

(b)

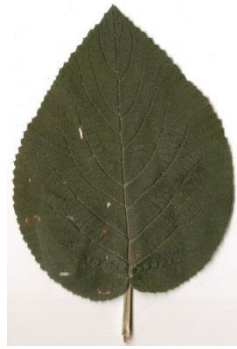
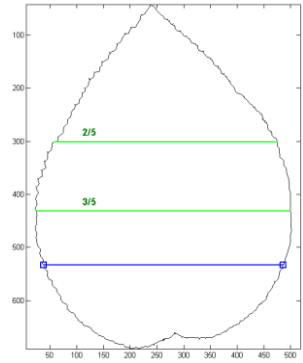

(c)

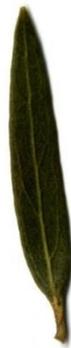
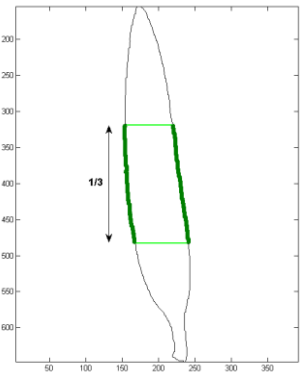

(d)

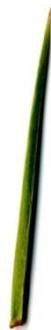
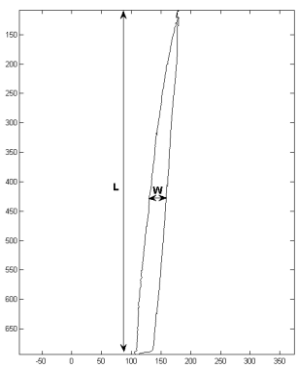

(e)

**Figure 5** Laminar shape classification (a) Elliptic (b) Obovate (c) Ovate (d) Oblong (e) Linear

### 3.3. Medial symmetry

The medial symmetry is determined by the ratio of the widths of the two halves of the leaf, separated with respect to the midvein in the middle of the leaf (from $0.25L$ to $0.75L$). This ratio is depicted in Fig.6 as X/Y (X being the lesser of the two distances).

Based on this character, the leaves are quantitatively categorized into two classes:

(1) *Symmetrical* - (X/Y) > 0.9.

(2) *Asymmetrical* - (X/Y) < 0.9.

As seen in the Fig.6, identifying the leaf's midvein is needed for finding the width ratio. But, as stated before, identifying the vein pattern from just the adaxial image capture is difficult and erroneous. Instead, we use the same estimate described in 2.5 (blue line in Fig.6). It should be noted that this is based on the assumption that the midvein is straight and devoid of any significant curves (which holds true in most cases).

The distance from the leaf boundary to this line is calculated in both sides from $0.25L$ to $0.75L$ and the ratio is computed. Then, we average the computed ratios to get a statistical estimation of the medial symmetry (Fig.6).

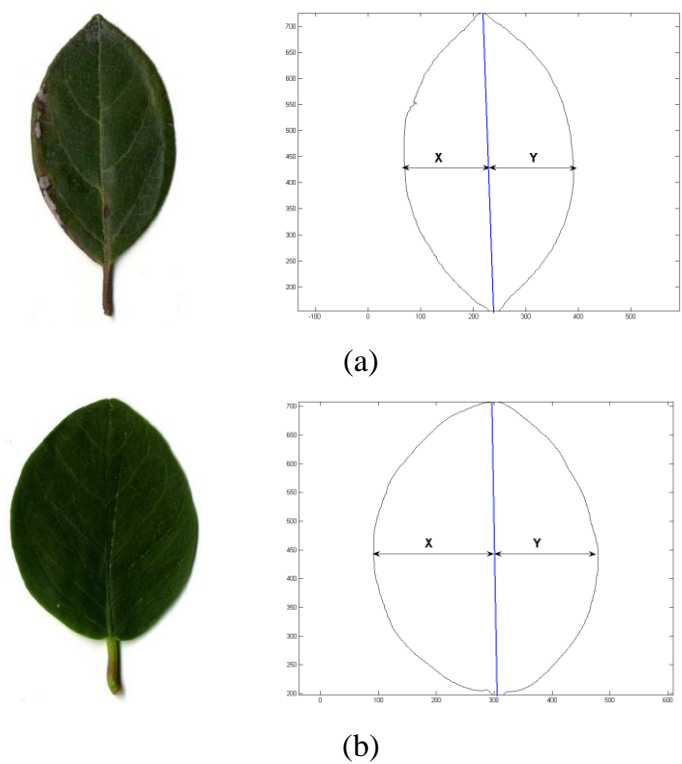

**Figure 6** Leaf medial symmetry (a) Symmetrical (b) Asymmetrical

## 3.5. Lobation

One of the most prominent and important features in describing a leaf's architecture is *lobation*. A *lobe* is defined as "a marginal projection with a corresponding sinus incised 25% or more of the distance from the projection apex to the midvein, measured parallel to the axis of symmetry" [16]. The leaves are categorized into two main classes: (1) Unlobed and (2) Lobed. Additionally, lobed leaves are then divided into multiple subclasses based on the type of lobation [16]. But since some of these types can only be identified by exploiting the vein pattern, we instead use the number of lobes which can be extracted from the leaf's shape only.

Our approach to lobe identification benefits from geometric characteristics of the lobed leaves. It has two main steps:
1) Separating the leaves which can be classed as *definitely-unlobed* (unlobed with 100% certainty).
2) Analyzing the remaining leaves which *probably* have lobes and if lobed, counting the number of them.

Since lobation is considered a large-scale change in leaf's margin, it is useful to employ an algorithm to convert the leaf's boundary to the simplest polygon possible (with as little vertices as possible), hence only reserving the basic outline of the leaf and eliminating minor irregularities. For this purpose, an algorithm called *Polygon Decimation (Simplification)* is applied (developed based on the surface simplification algorithm presented in [20]). Essentially, it will only keep important vertices from the original polygon and remove the ones that cause minimum error. It can be summarized in the steps below:

**Step1:** For every vertex compute the boundary offset error.
**Step2:** Rank all vertices according to the error score from step 1.
**Step3:** Remove the vertex with the lowest error.
**Step4:** Re-compute and accumulate the errors for the two neighbors adjacent to the deleted vertex and go back to step 2.
**Step5:** Repeat step 2 to 4 until no more vertices can be removed or the number of vertices has reached the desired number.

The polygon decimation algorithm is given the fraction of the points to reserve from the original polygon (denoted by *retain percent*). As we need minimum vertices for categorizing leaves into lobed or unrobed, the retain percent would be set **1%**. In other words, only one percent of the leaf boundary points is sufficient to identify the *definitely-unlobed* leaves. Fig.8 (a) shows the original and the decimated boundary of a lobed leaf (displayed in blue, denoted by $P_d$). It should be mentioned that *Ramer–Douglas–Peucker* algorithm [21, 22] could also be used for polygon simplification.

Another polygon used in this step is what is often referred to as *Convex Hull*. The Convex Hull or "Convex Envelope" of a set *S* of points in the Euclidean space is the smallest convex set that contains *S*. The Convex Hull can be visualized as the shape formed by a rubber band stretched around a set of nails. Here, the well-known *QuickHull* algorithm [23] is used to find the Convex Hull around the simplified polygon (in Fig.8 (c) the Convex hull is displayed in red, denoted by $P_{ch}$).

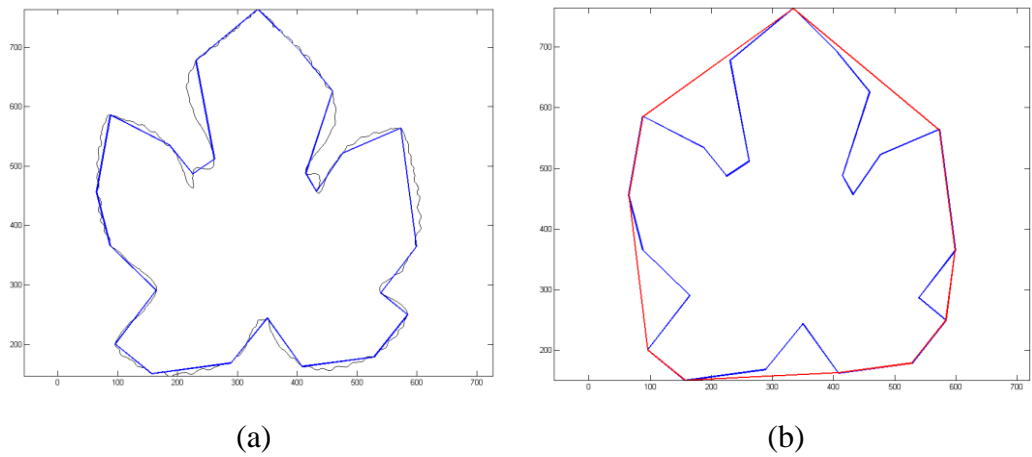

(a)          (b)

**Figure 7** Polygon decimation and convex hull of a lobed leaf (a) Reduced polygon - $P_d$ (b) Convex hull of $P_d$ (denoted by $P_{ch}$)

As the lobation is the main source of concavity in the leaf boundary (and subsequently in its reduced polygon representation), it is apparently the cause of the difference between polygons in Fig.8 (c). Thus, by measuring this difference, the unlobed leaves can be easily detected. For this measurement we use the polygons' areas (calculated by polygon triangulation). The more the leaf's lobation, the more area difference will occur between the decimated polygon and its convex hull. If this difference is below a threshold (set as **5%**), we identify that leaf as *definitely-unlobed*. Above this threshold and the leaf is considered probably lobed:

$$Lobation: \begin{cases} \frac{Area(P_{ch}) - Area(P_d)}{Area(P_{ch})} < 5\% : \; definitely\ unlobed \\ otherwise \qquad\qquad\qquad\qquad\ : \; probably\ lobed \end{cases}$$

Although we could identify the latter as (definitely) lobed leaves with an acceptable error margin, we postpone their final categorization after the second step to keep the error as minimum as possible. Fig.9 demonstrates a lobed and an unlobed leaf and their respective area difference (colored area).

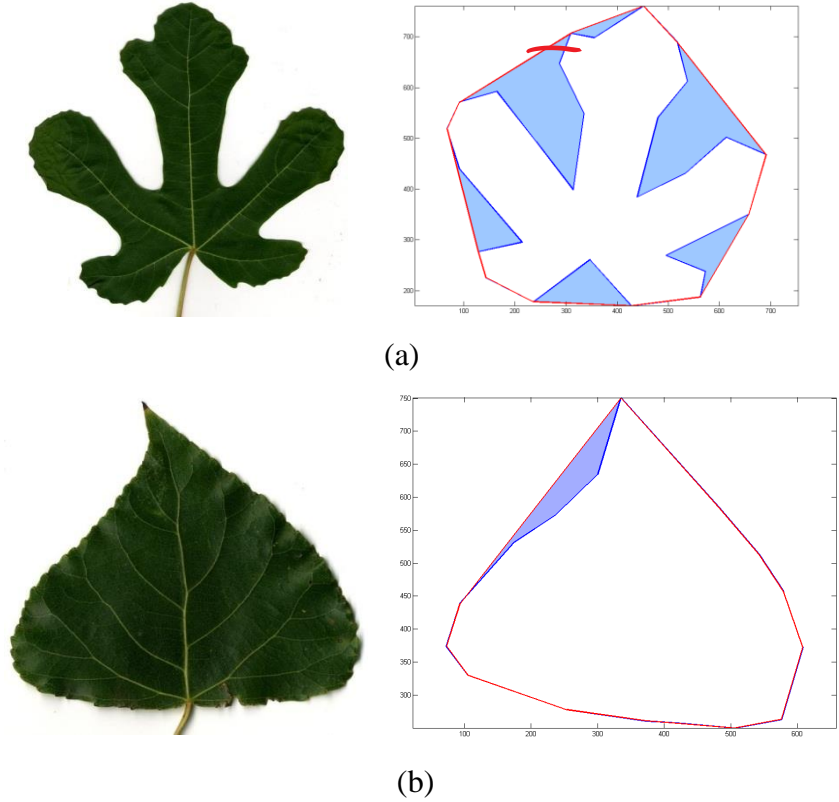

(a)

(b)

**Figure 8** the area difference of (a) a lobed leaf (42.60% area difference) (b) an unlobed leaf (4.97% area difference)

For step 2, again we analyze the relation between the decimated polygon and its convex hull ($P_{ch}$). As can be seen in Fig.9 (a), the points incised from the margin will not meet the convex hull and the apex points of the lobes are likely to be included in the convex hull boundary. We use this character to separate the incision zones from lobe apexes. Assuming the convex hull joins the $P_d$ at $k$ disjoint positions (in case of Fig.9 (a), $k=5$), the vertices of $P_d$ (denoted by $V_d$) can be divided into $k$ sets of incision vertices ($VI$) and $k$ sets of bound vertices ($VB$). Thus we have:

$$V_d = VI \cup VB$$

$$VI = \bigcup_{j=1}^{k} VI_j$$

$$VB = \bigcup_{j=1}^{k} VB_j \qquad (3)$$

For accurately counting the number of lobes, only one point from each $VI_j$ and one point from $VB_j$ should be counted. These will make $2k$ points that represent the incision vertices and apex vertices of *k* lobes. The best selection these points is the most *inner* points of incisions and the most *outer* points as apexes.

To perform this selection, we use the points' distance to the *centroid* of the leaf (marked by black point inside the leaf in Fig.10). The centroid of a 2-D shape is mathematically defined as the arithmetic average position of all the points inside the shape. The distance to the centroid is the principle component in Centroid Contour Distance (CCD), which is a popular method for identifying leaf lobes [5, 14]. By calculating the distance to centroid, we select the *nearest point to the centroid* from each $VI_j$ and *furthest point from the centroid* from each $VB_j$ (marked by blue and red squares respectively in Fig.10 (a)).

In [16] it is indicated that the incision should be more than *25%* of the projection peek's distance to the midvein to be considered a lobe. This quantitative measure is also applied to filter the incised points selected previously. The petiole insertion point (if *reflex*) is also filtered out. Finally, the actual number of lobes will be counted as *(number of incision points + 1)*. Fig.10 (b) shows the result of the method in a leaf with 5 lobes.

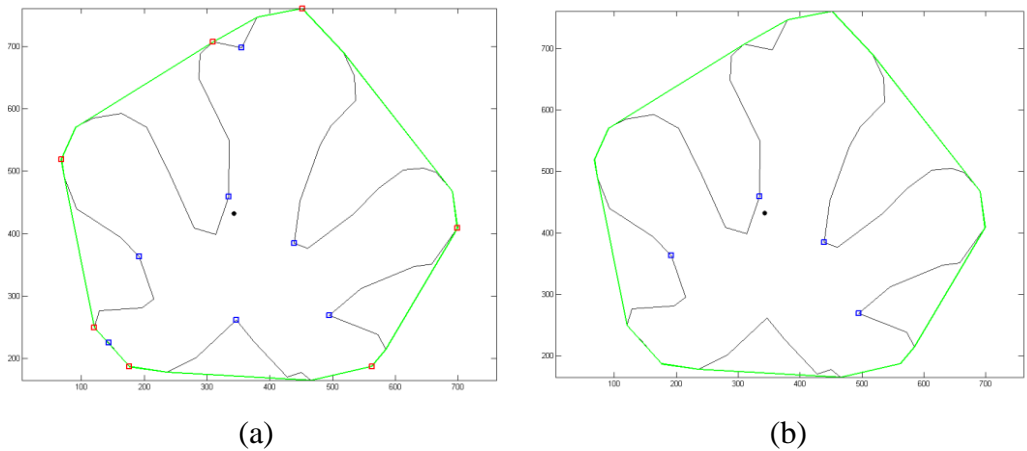

(a)          (b)

**Figure 9** Identifying and counting the lobes (a) All the incision and bound vertices (b) remaining incision vertices after filtering

### 3.6. Margin Type

Margin type or the tooth type of a leaf is one of the most difficult features to identify from the image mainly due to its small scale. On a higher level, leaves can be categorized as *toothed* or *untoothed*. The toothed leaves are then categorized into 3 types, *Dentate*, *Serrate* and *Crenate* (illustrated in Fig.11):

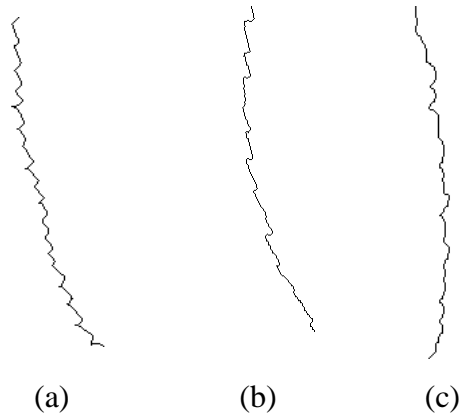

**Figure 10** Three types of teeth (a) Dentate (b) Serrate (c) Crenate

As can be seen in Fig.11, detecting these 3 types is difficult and highly prone to error. Thus, at this step we only categorize leaves into toothed and untoothed.

For this, we propose using a polygon's **perimeter** and its change in respect to the change in polygon **area** as a property that differentiates a toothed leaf from an untoothed one. It is geometrically evident that a toothed leaf has a greater perimeter than an untoothed leaf with the same polygon area. Applying this fact, we continuously decimate the boundary (discussed in section 2.5) while keeping the difference in area under 1%, until we find the most reduced polygon possible. When found, we observe the change in the perimeter. If the perimeter difference to the original polygon is small (less than *10%*), we consider the leaf as *toothed*. Otherwise, it is classified as *untoothed*.

**3.7. Apex angle and shape**

The apex angle is defined as "the angle between the two lines departing the apex vertex and tangent to the margin on each side of the leaf" [16]. In toothed leaves, they go cross the tips of the teeth. In lobed leaves, only the terminal lobe is taken into consideration. Based on this angle, leaves are categorized into 3 classes:

(1) *Acute* – apex angle < 90°

(2) *Obtuse* – apex angle between 90° and 180°

(3) *Reflex* – apex angle > 180°

The reflex apex is happened when the midvein ends at a bottom of a notch at the top of the leaf. The *reflex* apexes are identified in the 3rd stage of the preprocessing (section 2.3), so we don't repeat the process here. For calculating the apex angle, the convex hull of the apex part (distal ~25% of the lamina) is built and on each side of the apex vertex, the longest edge of the convex hull polygon ($P_{ch}$) is found. The angle between these two lines forms the apex angle and determines the type of the leaf. Fig.14 illustrates 3 leaves identified as each of the types above.

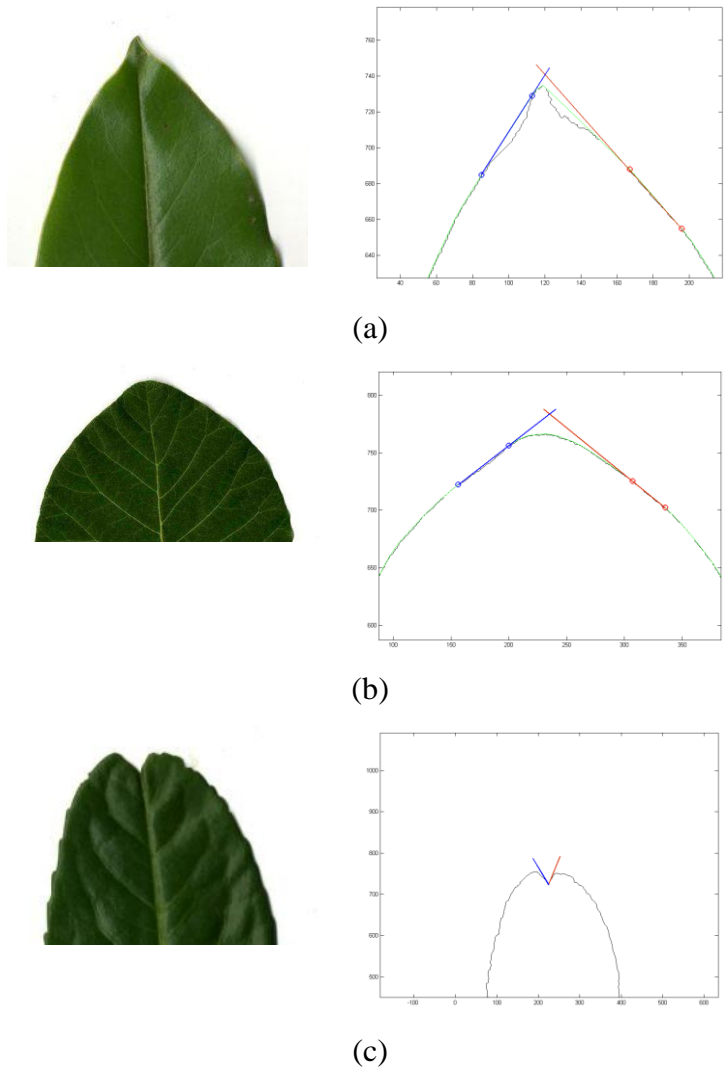

(a)

(b)

(c)

**Figure 11** Apex angle classification (a) Acute (b) Obtuse (c) Reflex

The apex based on its shape can be of 4 main types:
(1) *Straight* – The margin of the apex has no significant curvature.
(2) *Convex* – The margin of the apex curves away from midvein.

(3) *Acuminate* – The margin of the apex is convex proximally and concave distally, or concave only.

(4) *Apex with extension* – the apex vertex is extended downward.

The fourth type can be automatically identified as extended apexes have *reflex* apex angles (Fig.14 (c)) which are found in the preprocessing section (section 2.3). For identifying the first 3 types, we perform analysis on **numerical second derivative** of the apex contour. But, since the teeth may hinder with calculation, a principle shape of the apex should first be formed. For this purpose, the **Elliptic Fourier Transform (EFD)** [24] is employed. Essentially, FED represents a closed contour shape and coupled with its reverse transform using reduced harmonics, permits averages the shape of the leaf leaving the unwanted details (i.e. teeth) out. By reconstructing the shape of the apex using 12 harmonics and 1/5 of the original number of point, an averaged apex of the leaf is obtained (Fig.15).

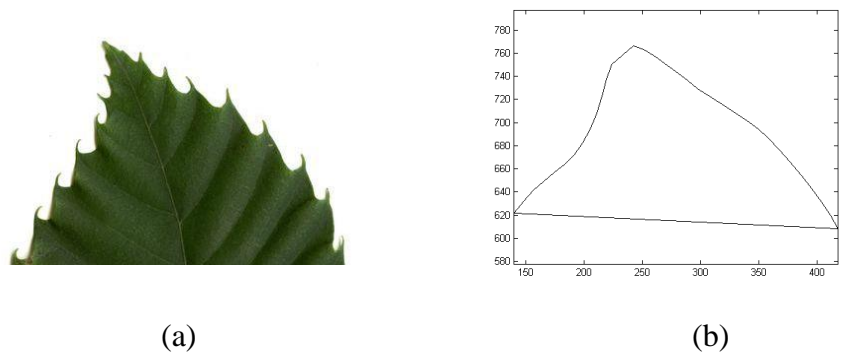

(a)          (b)

**Figure 12** Leaf apex averaging using FED (a) Original apex of the leaf image (b) reconstructed apex

By calculating the numerical second derivative of the apex vertices (*V*), we get an insight of how the curvature changes. Vertices at which the second derivative is positive ($V_+$) shows upward curvature at that location and points with negative second derivatives ($V_-$) display downward curve. For the categories aforementioned, we calculate the ratio of $\frac{V_+}{V}$ (in percent, called **Positive Second Derivative**) and apply the following rule:

$$Apex\ Shape : \begin{cases} PSD > 50\% & : Acuminate \\ 30\% < PSD < 50\% & : Straight \\ PSD < 30\% & : Convex \end{cases}$$

Fig.16 shows leaves categorized. into each of these types along their PSD values.

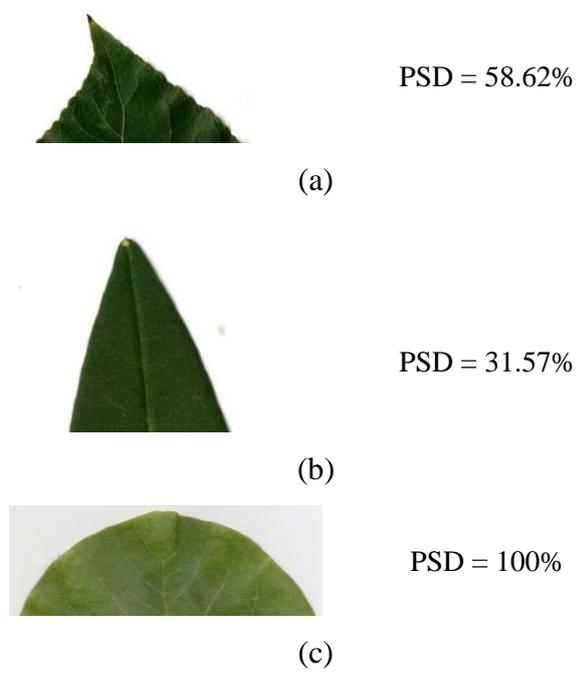

PSD = 58.62%

(a)

PSD = 31.57%

(b)

PSD = 100%

(c)

**Figure 13** Apex shape classification (a) Acuminate (b) Straight (c) Convex

## 3.8. Base angle and shape

The base angle is the angle between the two lines departing the base vertex and are tangent to the margin on each side of the leaf. Based on this angle, leaves are sorted into 3 types (identical to apex angle types, with the same angle thresholds): *Acuminate*, *Obtuse* and *Reflex*.

The *reflex* base type was identified during base symmetry check (section 2.4). Here, we focus on categorizing the other two types. Similar to the approach used in finding the apex angle, the convex hull of the base part (basal ~25% of the lamina) and the longest edge on each side is found. The angle between these lines forms the base angle (Fig.17).

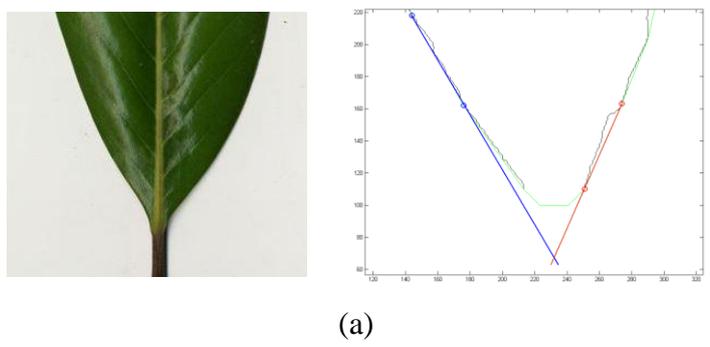

(a)

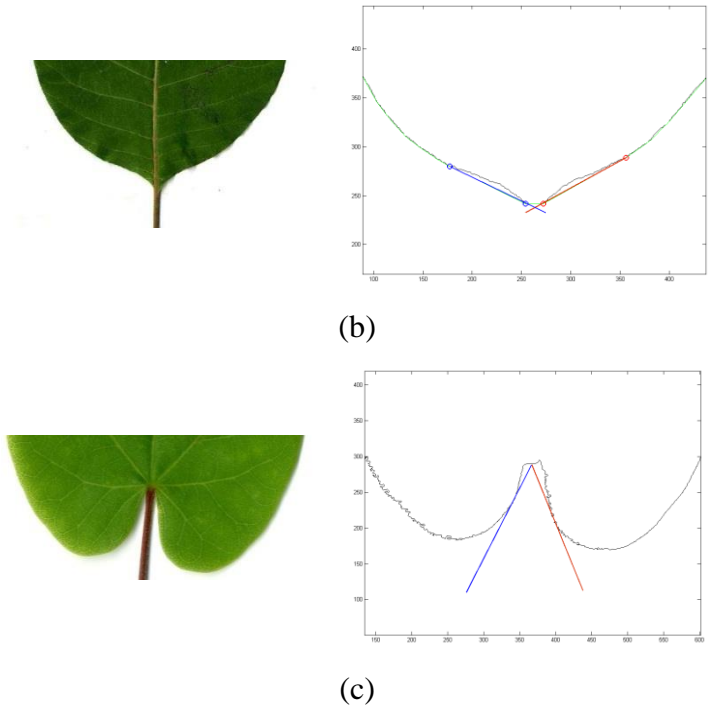

(b)

(c)

**Figure 14** Base angle classification (a) Acute (b) Obtuse (c) Reflex

Depending on the state of the *basal extension* (identified in section 2), the base shape is sorted into two groups:

- Without basal extension:

    (1) *Straight* – Base margin has no significant curvature.

    (2) *Concave* – Base margin curves toward the midvein.

    (3) *Convex* – Base margin curves away from the midvein.

    (4) *Concavo-convex* – Base margin is concave proximally and convex distally.

    (5) *Complex* – Margin curvature has more than one inflection point.

- With basal extension:

    (1) *Cordate* – The two sides of the petiole form sinus curves away from the petiole.

    (2) *Lobate* – Leaf base is lobed on both sides of the midvein.

(It should be noted that there is a sixth type in the first group, referred to as *Decurrent*, which is highly dependent on the exact trend of the midvein and hence, is not considered in this paper. We simply categorize these leaves as *concave*.)

For categorization in the first group, similar to finding the apex shape but with **Negative Second Derivative**, the $NSD = \frac{V_-}{V}$ (in percent) is calculated and the leaf is assorted by this rule:

$$Apex\ Shape : \begin{cases} NSD > 20\% & : Concave \\ 10\% < NSD < 20\% & : Straight \\ NSD < 10\% & : Convex \end{cases}$$

Now, because of the possible error and ambiguity in finding the exact position of the insertion point of the petiole, the 4$^{th}$ and 5$^{th}$ types are too close to other types to be identified successfully. Thus, we only take the first three types into account. Fig.18 shows three leaves categorized into the above classes and their according NSD's.

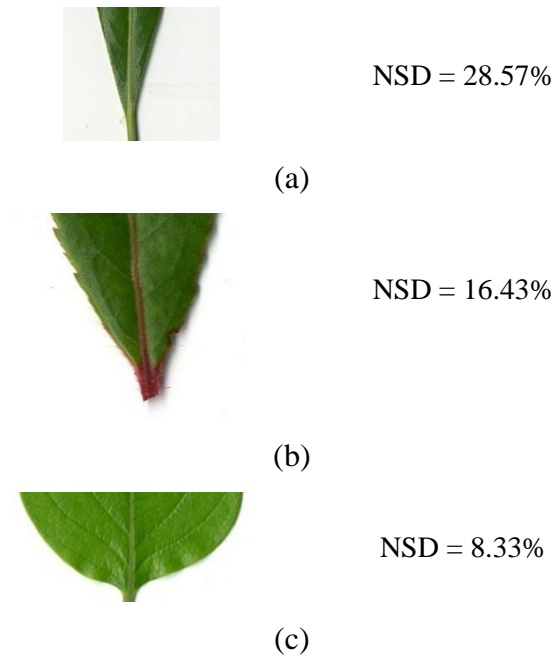

NSD = 28.57%

(a)

NSD = 16.43%

(b)

NSD = 8.33%

(c)

**Figure 15** Base shape classification (a) Concave (b) Straight (c) Convex

Regarding the second group, as we have calculated the apexes of the lobes (section 2.5), we apply the following rule: if at least one lobe apex is included in the base region, the leaf is *lobate*. Otherwise, it is categorized as *Cordate* (Fig.19).

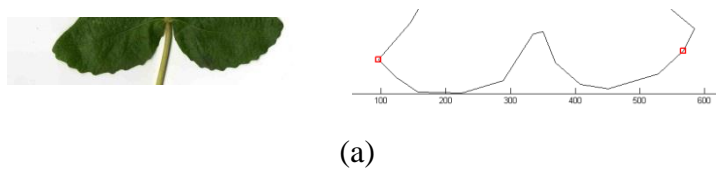

(a)

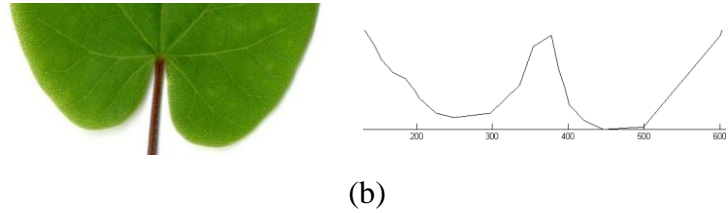
(b)

**Figure 16** Classification of extended bases (a) Lobate (b) Cordate

## 4. Classification and Species Identification

For classifying images based on the extracted features, we use **Naïve Bayes classifier**. The reasons for choosing this algorithm over others like Nearest Neighbors or Neural Networks include:

1) It is suitable in a problem space with many dimensions (here, 10).

2) We want the output to be a list of most probable species. This algorithm works well in these situations as it assigns a probability to each candidate.

3) The features used in this research are discrete, so no distance between their values can be defined. Thus, algorithms like K-NN are not suitable options.

4) Since allocating different weighs to the features should be done by an expert in botany or taxonomy, we weigh all features equally. This approach is convenient for Naïve Bayes classifier, but not for algorithms such as decision trees.

Of course, it should be mentioned that independency between features is one of the principle assumptions in Naïve Bayes classifier. But this classifier is also used with efficiency in contexts that this assumption doesn't hold (like text classification [?]). In our research, some minor dependencies among features are expected. Nevertheless, it shouldn't have significant effect on the performance of the algorithm.

In Naïve Bayes classification, considering n features, a new sample is classified by this formula:

$$Classify(f_1, f_2, \ldots, f_n) = \operatorname*{argmax}_{c} p(C = c) \prod_{i=1}^{n} p(F_i = f_i | C = c)$$

Here, $f_1$ to $f_n$ are the values of a new sample's features, $p(C = c)$ the probability of each class and $p(F_i = f_i|C = c)$ is the probability of the value $f_i$ for the feature $F_i$ given the sample belonging to the class C.

As explained before, we presume equal probability for the classes (species). Therefore, $p(C = c)$ would be $\frac{1}{G}$ with G as the number of species in our recognition system.

The probability of occurrence for each feature, as they are discrete, is estimated using $p = \frac{n_c}{n}$ with $n_c$ the number of its occurrences and $n$ total number of samples. Hence, the probability of each value, $p(F_i = f_i|C = c)$, can be found by calculating the number of times a value of $f_i$ is occurred for the feature $F_i$ and dividing it by the number of samples in species C.

Based on the above method, for each leaf sample, we calculate the probability of it belonging to any species in the system. Then, the top 3 most probable species are chosen as output.

## 5. Results

The proposed method in this paper was tested in two phases. In the first phase we test the features that don't require special measurements and can be checked simply by looking at the pictures. These features are *leaf organization, lobation (lobed/unlobed), margin type (toothed/untoothed), apex angle (reflex/non-reflex), base angle (reflex/non reflex)*.

Each of the above features would split a set of leaves into two classes (binomial classification) and therefore be greatly useful in certain classification methods like decision trees.

We use images from ImageCLEF 2012 dataset [25] to test the for these features. It is a dataset containing about 10,000 leaf images along with their taxonomy. Some examples of the images included in this dataset are shown in Fig.20.

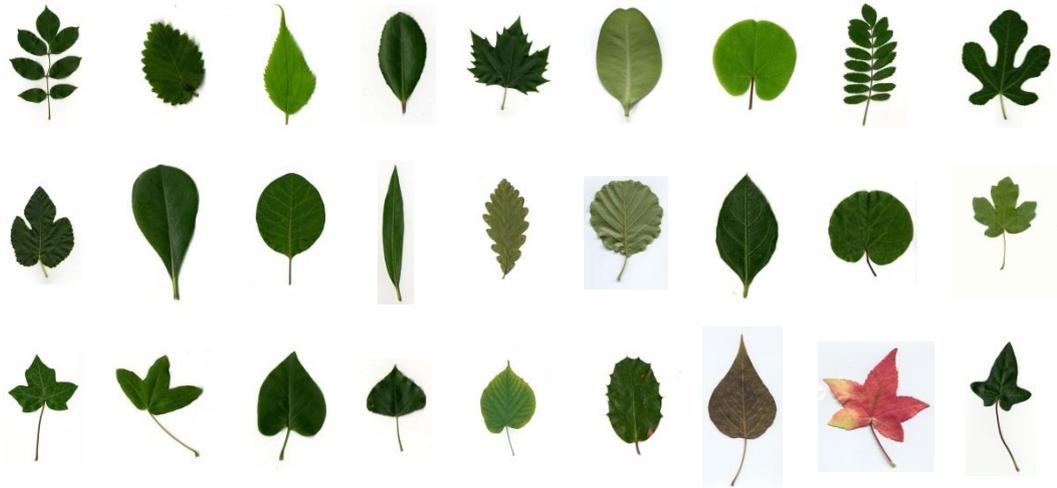

**Figure 17** Examples of the leaves in ImageCLEF 2012 dataset

For testing each of the features, two sets of leaf images are selected from the dataset: one set consisting of images known to be positive for the feature (e.g. a set of compound leaves) and the other known to be negative (with *PIM* and *NIM* denoting the number of images in each set). Testing the methods on these sets yields false or true positives (FP and TP) and false or true negatives (FN and TN). Using these rates, we then measure *precision* ($= \frac{TP}{TP+FP}$) and *recall* ($= \frac{TP}{TP+FN}$) for each feature. Also, as a final score on the performance of the approach on the specified features, the *F-measure* (*$F_1$ score*) defined as $F_1 = 2 \times \frac{percision \times recall}{percision + recall}$ is computed. $F_1$ shows the accuracy of the methods, reaching its best score at 1 (100% accurate) and worst at 0 (No accuracy). The results are displayed in Table.1.

**Table 1** Results of phase 1

| Feature | PIM/NIM | TP | FP | TN | FN | Precision | Recall | $F_1$ |
|---|---|---|---|---|---|---|---|---|
| Leaf Organization | 200/200 | 0.96 | 0.03 | 0.97 | 0.03 | 0.969 | 0.965 | 0.967 |
| Lobation | 100/100 | 1.00 | 0.07 | 0.93 | 0.00 | 0.934 | 1.000 | 0.966 |
| Margin Type | 100/100 | 0.87 | 0.1 | 0.90 | 0.13 | 0.897 | 0.870 | 0.883 |
| Apex Angle | 20/100 | 0.60 | 0.18 | 0.82 | 0.40 | 0.820 | 0.600 | 0.693 |
| Base Angle | 50/50 | 0.94 | 0.02 | 0.98 | 0.06 | 0.979 | 0.940 | 0.959 |

As it can be seen in Table.2, the proposed methods yield excellent results in binary classification on tested features.

In phase 2, we use all features and test the classification accuracy of our proposed method. Again, we use the ImageCLEF 2012 dataset for this purpose. This dataset includes images of 77 species in three types: scanned images, semi-scanned (photographed with controlled conditions) and photographed. Among all, only 38 species images consist of non-compound leaves on which our proposed algorithm can be applied to. And we used scanned and semi-scanned images as only they have the proper conditions for our proposed feature extraction techniques. In our implementation, 2/3 of the images were put for training and 1/3 for testing (randomly chosen).

The average recognition rate of our method on these images was **58.3%**, considering only the first output (top probability). If we take into account all three outputs of the algorithm, the accuracy would go upper. In comparison, one of the contestant teams in ImageCLEF 2012 that similarly used leaf features for classification, reported 43% recognition on scanned images which put them on second place. Of course, their input was all 77 species, but also the training and test sets were fixed for them.

## 6. Conclusion and future work

In this paper an approach for extraction of architectural features of plant leaves has been introduced. Also, for each feature, the quantitative rules for classification based on standard botanical definitions were presented, which can be used for automated leaf classification. Primary tests show promising results in case of binary classification of random leaves.

Although, high error rate emerged when testing was done on a standard dataset (ImageCLEF 2012). Some reasons for this error can be:
1) Different thresholds were presumed in our method which can cause high error if not set properly.

2) Inherent error of employing only the botanical features of a leaf lamina.

3) Error in preprocessing due to different conditions of the leaf shape and the image itself.

4) Insufficient number of features for successfully classifying a leaf based on its image.

In future, this work can be used as a framework for more detailed and more accurate classifications, combining high-level linguistic features presented here with low-level computer vision features in plant identification systems. Additionally, the proposed approach can be employed by botanists to statistically gather high-level classification information on images of leaves in the same family, genus and species. Hence, it becomes possible to study the distribution of extracted feature values and classes inside the mentioned categories of plants.